\documentclass[conference]{IEEEtran}

\IEEEoverridecommandlockouts

\usepackage{cite}
\usepackage{amsthm}  
\usepackage{amsmath, amssymb, amsfonts}  
\usepackage{algorithmic}
\usepackage{graphicx}
\usepackage{textcomp}
\usepackage{booktabs}
\usepackage[table]{xcolor}  
\usepackage{mathtools}
\usepackage{microtype}
\usepackage{subfigure}
\usepackage[colorlinks]{hyperref}
\usepackage{float}

\def\BibTeX{{\rm B\kern-.05em{\sc i\kern-.025em b}\kern-.08em
    T\kern-.1667em\lower.7ex\hbox{E}\kern-.125emX}}
\begin{document}

\title{DA-Mamba: Domain Adaptive Hybrid Mamba-Transformer Based One-Stage Object Detection
}

\author{\IEEEauthorblockN{1\textsuperscript{st} A. Enes Doruk}
\IEEEauthorblockA{\textit{Department of Artificial Intelligence and Data Eng.} \\
\textit{Ozyegin University}\\
Istanbul, Turkiye \\
enes.doruk@ozu.edu.tr}
\and
\IEEEauthorblockN{2\textsuperscript{nd} Hasan F. Ates}
\IEEEauthorblockA{\textit{Department of Artificial Intelligence and Data Eng.} \\
\textit{Ozyegin University}\\
Istanbul, Turkiye \\
hasan.ates@ozyegin.edu.tr}
}

\maketitle

\begin{abstract}

Recent 2D CNN-based domain adaptation approaches struggle with long-range dependencies due to limited receptive fields, making it difficult to adapt to target domains with significant spatial distribution changes. While transformer-based domain adaptation methods better capture distant relationships through self-attention mechanisms that facilitate more effective cross-domain feature alignment, their quadratic computational complexity makes practical deployment challenging for object detection tasks across diverse domains. Inspired by the global modeling and linear computation complexity of the Mamba architecture, we present the first domain-adaptive Mamba-based one-stage object detection model, termed DA-Mamba. Specifically, we combine Mamba’s efficient state-space modeling with attention mechanisms to address domain-specific spatial and channel-wise variations. Our design leverages domain-adaptive spatial and channel-wise scanning within the Mamba block to extract highly transferable representations for efficient sequential processing, while cross-attention modules generate long-range, mixed-domain spatial features to enable robust soft alignment across domains. Besides, motivated by the observation that hybrid architectures introduce feature noise in domain adaptation tasks, we propose an entropy-based knowledge distillation framework with margin ReLU, which adaptively refines multi-level representations by suppressing irrelevant activations and aligning uncertainty across source and target domains. Finally, to prevent overfitting caused by the mixed-up features generated through cross-attention mechanisms, we propose entropy-driven gating attention with random perturbations that simultaneously refine target features and enhance model generalization. Extensive experiments demonstrate that DA-Mamba consistently outperforms existing methods across a range of widely recognized domain adaptation benchmarks. our code available at \href{https://github.com/enesdoruk/DAMamba}{enesdoruk/DAMamba}.
\end{abstract}


\begin{figure}[!ht]
\centering
\includegraphics[width=1 \linewidth]{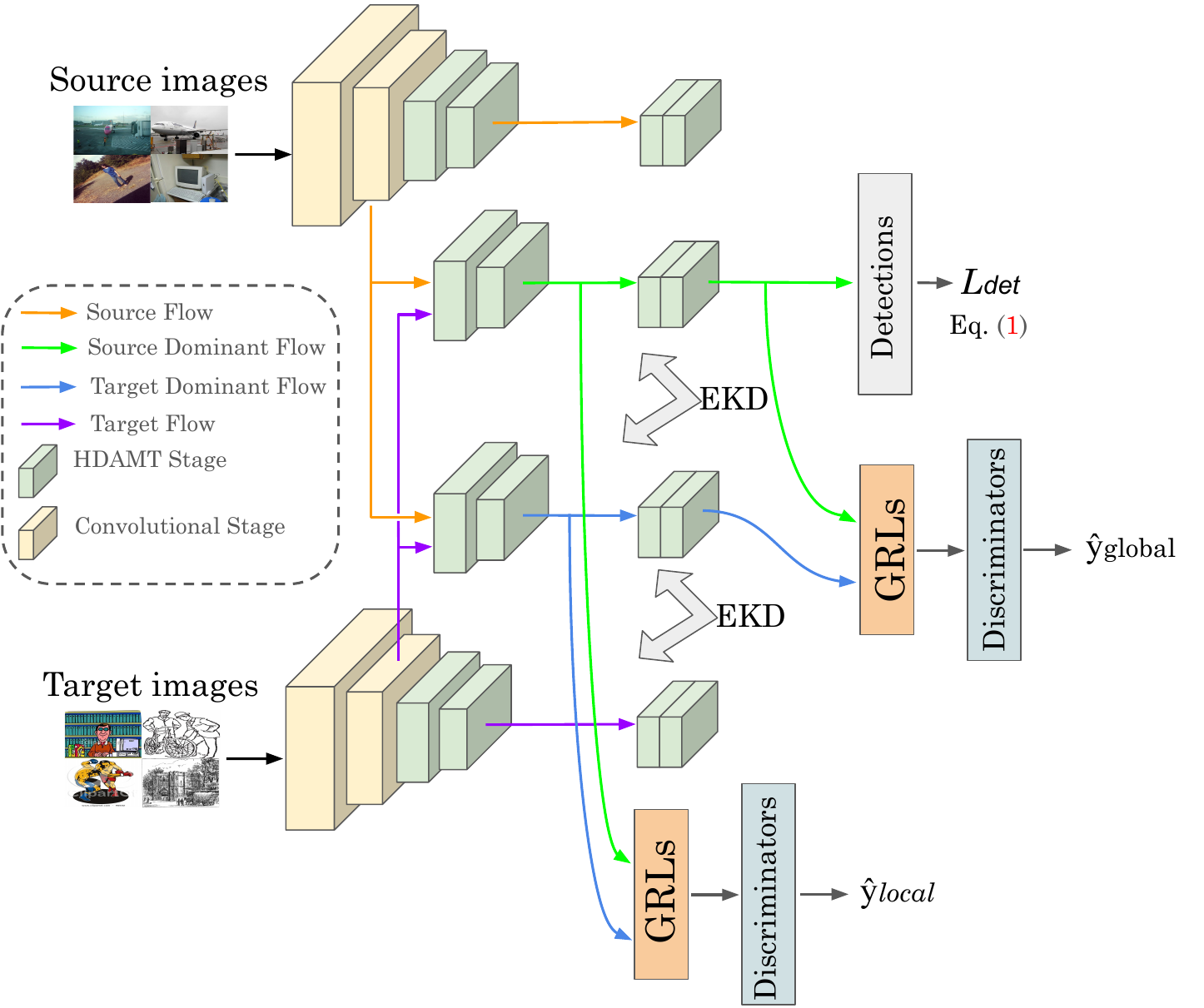}
\vskip-4.5pt
\caption{ Overall DA-Mamba architechture.}
\label{fig:arch}
\end{figure}

\section{Introduction}
\label{1}

Object detection identifies and localizes objects using bounding boxes and is widely used in autonomous driving, surveillance, robotics, etc. \cite{ren2016faster, liu2016ssd}. However, object detectors often struggle with domain shift, where performance degrades when tested on  different datasets. Unsupervised Domain Adaptation (UDA) mitigates domain shift by adapting models trained on labeled source domains to perform effectively on unlabeled target domains \cite{deng2021domain, chen2019progressive}. Key approaches include adversarial training with the Gradient Reversal Layer (GRL) \cite{ganin2015unsupervised} for image- and instance-level adaptation \cite{vs2021mega}, as well as pseudo-labeling strategies for refining target predictions \cite{vs2023instance}. 

Traditional 2D CNN-based object detectors face significant challenges in cross-domain adaptation scenarios due to their inherent architectural limitations~\cite{chen2020harmonizing, xu2020cross}. Both single-stage and two-stage detectors suffer from inadequate modeling of long-range dependencies because of their limited receptive fields, preventing effective relationship capture between distant image regions that are crucial for robust cross-domain feature alignment. Two-stage detectors like Faster R-CNN~\cite{ren2015faster}, while offering more comprehensive instance-level feature alignment through adversarial adaptation at local, global, and instance levels, are hampered by computationally expensive region proposal mechanisms and ROI-based operations that significantly reduce inference speed and complicate real-world deployment. Meanwhile, single-stage detectors such as SSD~\cite{liu2016ssd} and RetinaNet~\cite{lin2017focal} provide faster performance but lack explicit instance-level representations due to the absence of region proposal steps, making domain adaptation even more challenging. These limitations highlight the need for efficient architectures that capture long-range dependencies, motivating transformer-based or hybrid models that balance context modeling with speed.

Transformer-based object detection approaches like DETR~\cite{carion2020end} offer significant advantages for cross-domain adaptation by effectively addressing the long-range dependency limitations inherent in CNN-based detectors. Unlike conventional CNN architectures, DETR~\cite{carion2020end} utilizes self-attention mechanisms to capture global relationships across the entire feature space, modeling contextual information regardless of spatial distance. This approach eliminates hand-crafted components like anchors, non-maximum suppression, and region proposals. Research efforts such as DA-DETR~\cite{zhang2023detr} and DATR~\cite{chen2025datr} have explored leveraging DETR's attention mechanism for cross-domain feature alignment, using attention maps to identify domain-invariant regions and facilitate more effective knowledge transfer between source and target domains. However, transformer-based detectors face their own significant challenges for practical domain adaptation scenarios. Despite these benefits, the quadratic complexity of the attention mechanism with respect to sequence length makes Transformers computationally expensive to train and deploy. Additionally, cross-attention mechanisms often suffer from overfitting to source domain features, limiting their ability to generalize effectively to the target domain~\cite{alijani2024vision}. This overfitting arises as the model learns domain-specific correlations instead of capturing domain-invariant representations, reducing adaptability to unseen distributions. These computational and generalization bottlenecks present substantial barriers to their widespread adoption in resource-constrained real-world cross-domain applications where efficient and robust inference is crucial.

To address the challenges of limited long-range dependency modeling in CNNs, efficiency in hybrid CNN-Transformer architectures, and overfitting in transformer-based cross-domain adaptation, we propose DA-Mamba, the first Mamba-based network for domain adaptive object detection, achieved by redesigning a hybrid Mamba-Transformer~\cite{hatamizadeh2024mambavision} architecture for domain adaptation and integrating it into the SSD~\cite{liu2016ssd} framework. The overall architecture is illustrated in Figure~\ref{fig:arch}. Our domain-adaptive design employs Mamba blocks to perform adaptive feature scanning, enhancing domain-agnostic channel communication while counteracting spatial rigidity and suppressing negatively transferable features. To facilitate effective domain alignment, we utilize cross-attention modules that selectively extract source- and target-dominant features. Owing to the linear computational complexity of Mamba’s state-space models, DA-Mamba remains significantly more lightweight and efficient than conventional end-to-end Transformer architectures. Hybrid approaches often introduce noise in domain adaptation tasks due to the mismatch between the local feature extraction of CNNs and the global attention mechanisms of Transformers or Mamba, which can cause inconsistencies when transferring domain-specific features. To address this, we propose a novel Margin ReLU-guided entropy knowledge distillation framework, which promotes the bidirectional transfer of both source- and target-specific features while effectively suppressing noisy activations and enhancing domain alignment. Moreover, recognizing that cross-attention mechanisms often overfit to source-domain features and hinder generalization to target domains~\cite{alijani2024vision}, we develop an adaptive feature fusion module with an entropy-sensitive gating mechanism. By incorporating stochastic perturbations at multiple network depths, our approach improves robustness against domain shifts and mitigates overfitting, enabling more reliable and efficient cross-domain inference in resource-constrained environments. In summary, DA-Mamba offers long-range feature modeling with enhanced lightweight efficiency, effectively prevents overfitting through adaptive perturbations, and provides a deployment-friendly, single-stage domain adaptive detector suitable for real-world applications.

The key contributions of this paper are summarized as follows:
\begin{itemize}
        \item Hybrid Domain-Adaptive Mamba-Transformer integrates state-space models (SSMs) with attention mechanisms for robust feature alignment. The architecture scans transferable domain-specific features along both spatial and channel dimensions using Mamba blocks,  while deploying self-attention to enhance intra-domain representation. Simultaneously, cross-attention mechanisms extract source- and target-dominant features, facilitating soft alignment between domains for optimal transfer learning performance.
        
        \item We propose a novel Margin ReLU-guided entropy knowledge distillation mechanism that systematically facilitates bidirectional transfer of both source-dominant and target-dominant features. Recognizing the noise introduced by hybrid architectures in domain adaptation tasks, this approach strategically minimizes domain-specific distribution gaps while effectively suppressing noisy activations.
        
        \item  We develop an adaptive feature fusion framework with an entropy-sensitive gating mechanism that dynamically balances feature importance across domains. Our approach incorporates stochastic perturbations at multiple network depths, which enhances model robustness against domain variations and prevents overfitting to source-specific representations. 
\end{itemize}

\begin{figure*}[!t]
\centering
\includegraphics[width=0.75 \linewidth]{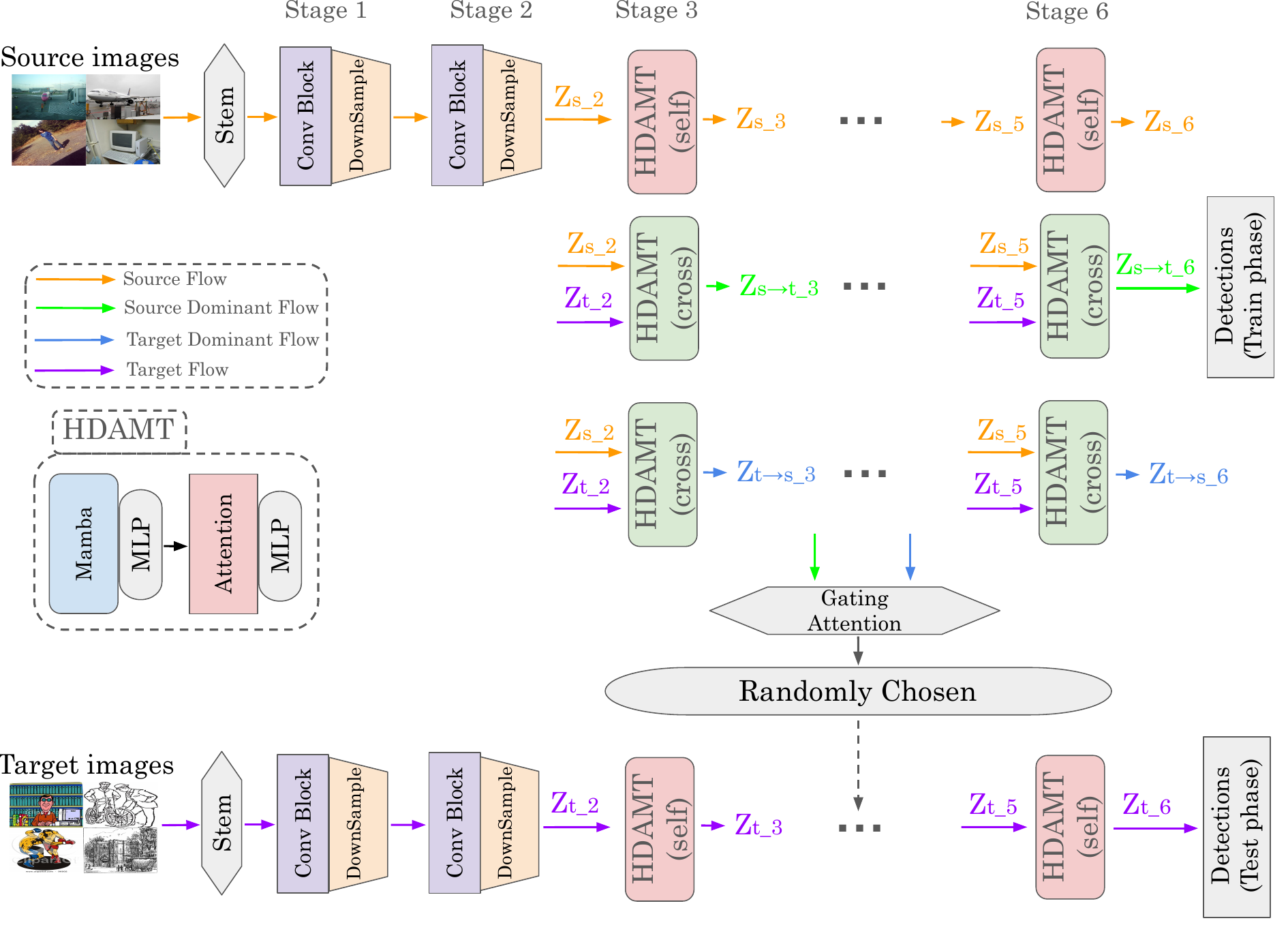}
\vskip-2.5pt
\caption{Overview of the Hybrid Domain-Adaptive Mamba-Transformer architecture, showing the flow of source, source-dominant, target-dominant, and target features across the stages.}
\label{fig:hdamt}
\end{figure*}

\section{Related Work}
\label{2}

\textbf{Domain Adaptive Object Detection.} Unsupervised domain adaptation (UDA) for object detection aims to mitigate domain gaps by training models on labeled source domains for deployment on unlabeled target domains. These methods primarily fall into three categories: adversarial training~\cite{ saito2019strong, zhu2019adapting}, self-training~\cite{wang2021domain}, and image-to-image translation~\cite{kim2019diversify}. Domain Adaptive Faster R-CNN~\cite{chen2018domain} pioneered this field by implementing adversarial feature alignment at both image and instance levels. Most existing adaptive detectors adopt two-stage CNN-based frameworks like Faster R-CNN~\cite{chen2021scale, zhuang2020ifan}, which rely heavily on region proposals for feature alignment. Although single-stage detectors are efficient in terms of inference, research on adapting them remains limited. These detectors face challenges such as limited receptive fields and inadequate modeling of long-range dependencies, which complicates domain adaptation. He et al.~\cite{he2019multi} implemented weak self-training by reducing false positives during hard negative mining, while Hsu et al.~\cite{hsu2020every} proposed progressive feature alignment for single-stage adaptation. Wang et al.~\cite{wang2021domain} introduced a teacher-student framework specifically for single-stage detectors to transfer domain knowledge. Despite these efforts, the fundamental challenge remains in learning instance-invariant feature representations without region proposal guidance. Unlike two-stage methods that can leverage proposal-based instance alignment~\cite{xu2020cross, zheng2020cross}, single-stage detectors struggle to capture fine-grained cross-domain correspondences due to their unified detection pipeline.

Transformer-based object detection approaches like DETR~\cite{carion2020end} have emerged as alternatives with strong global context modeling capabilities. Li et al.~\cite{li2022cross} proposed a domain-adaptive DETR framework that aligns cross-domain features through transformer attention mechanisms, while Zheng et al.~\cite{zheng2021implicit} introduced query-based alignment for domain transfer. Despite their effectiveness in modeling long-range dependencies, transformer-based adaptation methods face significant computational challenges. Their quadratic complexity with respect to sequence length~\cite{carion2020end} and memory-intensive training requirements make them less practical for resource-constrained deployment scenarios compared to optimized single-stage detectors. This efficiency gap highlights the need for adaptation approaches that can leverage global context modeling without the prohibitive computational costs associated with transformer architectures.

\textbf{Vision Mamba.} State Space Models (SSMs) offer a scalable alternative to attention mechanisms with linear time and space complexity. Mamba~\cite{gu2023mamba} advanced this line by introducing selective scanning and parallel processing, enabling efficient sequence modeling across modalities. Its application has since expanded into visual tasks through various adaptations. Vision Mamba~\cite{zhu2024vision} was the first to apply Mamba to computer vision, using Selective SSMs for a pure visual backbone. VMamba~\cite{liu2024vmamba} added a Cross-Scan Module (CSM) to enhance 2D directional awareness. LocalMamba~\cite{huang2024localmamba} adopted window-based dynamic scanning for local dependency modeling. MambaVision~\cite{hatamizadeh2024mambavision} further improved efficiency and accuracy with a streamlined forward pass and selective self-attention, surpassing Vision Mamba~\cite{zhu2024vision}. EfficientVMamba~\cite{pei2024efficientvmamba} integrated atrous convolutions and CNN-SSM hierarchies for global context, while MambaVision simplified this with a unified Mamba mixer. Recent extensions like MambaDETR~\cite{ning2024mambadetr}, and Mamba YOLO~\cite{wang2024mamba} show Mamba’s versatility in vision tasks. However, none address domain adaptation. Our work fills this gap with a Mamba-based approach for domain-adaptive visual learning.

\section{Method}
\label{3}

\begin{figure*}[!t]
\centering
\includegraphics[width=0.9 \linewidth]{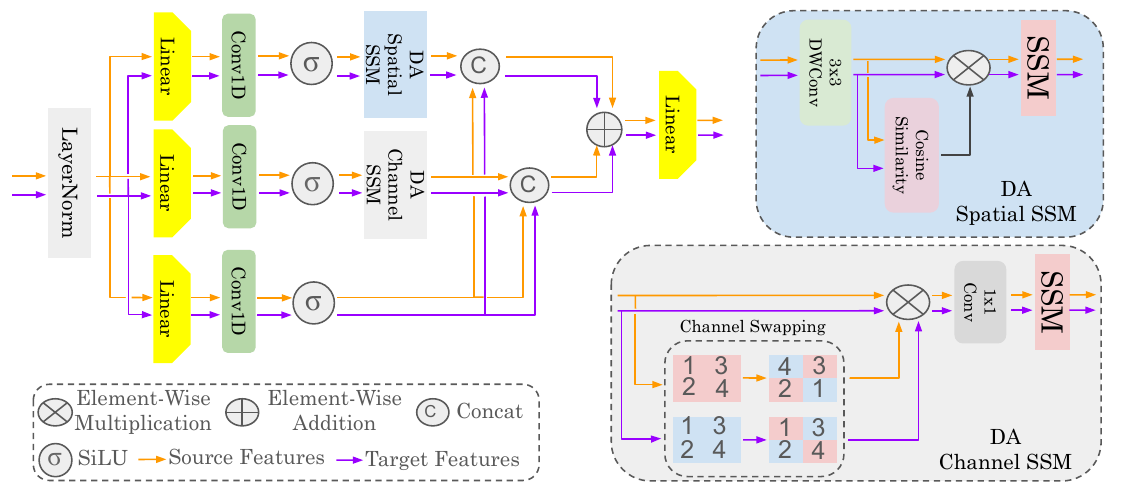}
\vskip-1.5pt
\caption{ Structure of the Mamba block, showing the Spatial SSM, Channel SSM.}
\label{fig:mamba}
\end{figure*}

\subsection{Task Definition}
\label{3_1}

The study addresses the challenge of unsupervised domain adaptation (UDA) for object detection. It considers a source domain $\mathcal{D}_s$ and a target domain $\mathcal{D}_t$, where $\mathcal{D}_s = \left\{ \left(x_s^{i}, y_s^{i} \right) \right\}_{i=1}^{N_s}$ is annotated, and $y_s^{i}$ denotes the set of ground-truth object instances for each image $x_s^{i}$, including bounding boxes and category labels.
The objective is to train a detector that generalizes well to the unlabeled target samples $x_t^i$. The initial model is trained solely on the labeled source dataset (i.e., $\mathcal{D}_s$):

\begin{equation} \label{eq:det}
L_{det}=l(Z(x_s), y_s),
\end{equation}
where $Z$ denotes the hybrid domain-adaptive Mamba-Transformer feature extractor, with $Z_{s \rightarrow t}$ representing the source-dominant features used during training, and $Z_t$ representing the target features used during inference (see Figure~\ref{fig:hdamt}). The supervised detection loss $l\left(\cdot\right)$ consists of classification and localization terms~\cite{liu2016ssd} for predicting object categories and bounding boxes.

\subsection{Framework Overview}
As illustrated in Figure~\ref{fig:hdamt}, we adopt a hybrid Mamba-Transformer backbone inspired by MambaVision~\cite{hatamizadeh2024mambavision} and integrate it into SSD~\cite{liu2016ssd} to address domain shift while enabling efficient long-range and domain-aware adaptation within a single-stage detection framework. Transferable representations are extracted via domain-adaptive spatial and channel-wise scanning through Mamba modules, enhancing geometric and semantic consistency. Self-attention mechanisms capture intra-domain context, while cross-attention generates source-dominant and target-dominant features ($Z_{s \rightarrow t}$ and $Z_{t \rightarrow s}$) for soft alignment, enabling the transfer of rigid, domain-specific features across domains. These features serve as the foundation for all subsequent adaptation modules.

To regularize alignment and suppress noise introduced by hybrid representations, we employ a Margin ReLU-guided entropy knowledge distillation (EKD) mechanism that enables structured bidirectional feature transfer using $Z_{s \rightarrow t}$ and $Z_{t \rightarrow s}$. To improve robustness, an entropy-based random multi-layer perturbation (ERMP) module also leverages these cross-attended features to adaptively perturb dominant intermediate and deep features, enhancing cross-domain generalization.

To reduce domain discrepancies in feature representations and improve generalization across domains, we perform adversarial alignment at both local and global levels using $Z_{s \rightarrow t}$ and $Z_{t \rightarrow s}$, as illustrated in Figure~\ref{fig:hdamt}. As shown in the discriminator block in Figure~\ref{fig:arch}, a pixel-level discriminator is applied to the third-stage HDAMT featuresfor local alignment and trained using cross-entropy loss, while a semantic-level discriminator is applied to the sixth-stage HDAMT features for global alignment and optimized using focal loss. The overall adversarial objective is defined as:
\begin{equation}
L_{\text{adv}} = \sum_{i} \text{CE}(\hat{y}_{\text{local}}^i, y_{\text{local}}^i) + \text{FL}(\hat{y}_{\text{global}}, y_{\text{global}}),
\label{eq:adv}
\end{equation}
Here, $\hat{y}_{\text{local}}$ and $\hat{y}_{\text{global}}$ denote the predicted domain labels, while $y_{\text{local}}$ and $y_{\text{global}}$ represent the corresponding binary ground-truth labels (source or target) used for supervision. We adopt focal loss to down-weight easy negative examples and emphasize hard, informative foreground instances, which is particularly effective for addressing semantic-level class imbalance in domain adaptation~\cite{lin2017focal}. Gradient Reversal Layers (GRLs)~\cite{ganin2015unsupervised} are inserted before each discriminator to reverse gradients and enforce domain confusion during training.

During training, we use the source-dominant features $Z_{s \rightarrow t}$ for detection to leverage stable supervision from the labeled source domain, while during inference we switch to $Z_{t}$ to ensure full adaptation to the target domain distribution.

\subsection{Hybrid Domain-Adaptive Mamba-Transformer}

The proposed Hybrid Domain-Adaptive Mamba-Transformer (HDAMT) backbone, illustrated in Figure~\ref{fig:hdamt}, adopts a six-stage architecture that balances local spatial encoding and global semantic alignment for cross-domain object detection. The first two stages consist of convolutional layers designed to capture low-level spatial patterns such as edges and textures, while the remaining four stages comprise hybrid Mamba-Transformer blocks that integrate domain-adaptive scanning and attention mechanisms. To support effective domain alignment, our framework constructs four feature branches during training: source features ($Z_s$), target features ($Z_t$), and cross-attention-derived source-dominant ($Z_{s \rightarrow t}$) and target-dominant ($Z_{t \rightarrow s}$) features. These cross-attended representations enable soft alignment by blending domain-specific and domain-agnostic cues.

Each Mamba-Transformer block, illustrated in Figure~\ref{fig:mamba}, follows a three-branch structure: (1) a spatial branch using the Domain-Adaptive Spatial Selective Scanning Mechanism (DA Spatial SSM), (2) a semantic branch using the Domain-Adaptive Channel Selective Scanning Mechanism (DA Channel SSM), and (3) a transformation branch. The transformation branch is concatenated separately with the outputs of both the DA Spatial SSM and DA Channel SSM. These two enriched representations are then fused via elementwise addition to form the final hybrid domain-adaptive feature representation.

\textbf{DA Spatial SSM}. It targets spatial alignment in early-to-mid stages where local structural patterns are prominent. Although Mamba layers use 1D convolutions, they lack spatial inductive bias crucial for vision tasks. To recover 2D spatial structure, we apply a 3×3 depthwise convolution after the 1D convolution on both source and target features. We then compute spatial cosine similarity and use it to reweight spatial locations, enhancing spatial consistency across domains.

\textbf{DA Channel SSM.} It addresses channel-level adaptation in deeper layers where semantic abstraction dominates. Feature channels are divided into four segments, and two are selectively swapped between the source and target domains. Channel swapping exposes each domain to the other's semantic features, encouraging the model to learn domain-agnostic correlations. A 1×1 convolution is then applied to recalibrate and fuse the interleaved channels.

Each hybrid block also includes self-attention and cross-attention. Self-attention models intra-domain dependencies, while cross-attention generates source- ($Z_{s \rightarrow t}$) and target-dominant ($Z_{t \rightarrow s}$) features to enable soft alignment between domains.

\subsection{Entropy-Based Knowledge Distillation}

To address the feature inconsistency and activation noise caused by mismatched spatial and semantic encoding in hybrid architectures, we propose an entropy-based knowledge distillation (EKD) module that aligns source- and target-specific features across multiple semantic levels. The EKD module uses cross-attention to extract source-dominant (\(Z_{t \rightarrow s}\)) and target-dominant (\(Z_{s \rightarrow t}\)) features, which serve as soft targets to guide the alignment between source (\(Z_s\)) and target (\(Z_t\)) representations.

Entropy is computed at three critical feature levels—shallow (stage 2), mid (stage 4), and deep (stage 6)—to supervise domain alignment hierarchically. Since source- and target-dominant features can introduce noise due to domain-specific biases, we apply a margin ReLU prior to entropy computation to suppress uncertain activations and prevent noisy feature contributions during alignment. The margin \(m\) is computed from batch normalization statistics, where each channel's threshold is estimated from the tail of a Gaussian distribution based on its activation mean and variance. This probabilistic thresholding adaptively filters out low-confidence or domain-specific responses while preserving discriminative signals.

The total entropy loss is formulated as:
\begin{equation}
\begin{aligned}
H_{t \rightarrow ts} &= -\sum_{k} \sigma_m({Z_{t \rightarrow s}}^{(k)})\log(\sigma_m({Z_t}^{(k)})), \\
H_{t \rightarrow st} &= -\sum_{k} \sigma_m({Z_t}^{(k)})\log(\sigma_m({Z_{s \rightarrow t}}^{(k)})),
\end{aligned}
\label{eq:entropy}
\end{equation}
where $\sigma_m$ denotes the margin ReLU, and $k$ indexes the feature channels. The final entropy loss is computed by averaging across all stages and samples:

\begin{equation}
L_{\text{entropy}} = \frac{1}{N} \sum_{l=2,4,6} \left( H_{t \rightarrow ts}^{(l)} + H_{t \rightarrow st}^{(l)} \right),
\label{eq:kd}
\end{equation}
where \(N\) is the batch size and \(l\) indexes the feature stage. This multi-scale supervision reduces uncertainty and improves domain-invariant representation learning.

\subsection{Entropy-Guided Random Multi-Layer Perturbation}
To address the tendency of cross-attention mechanisms to overfit to source-domain features and compromise generalization~\cite{alijani2024vision}, we introduce an entropy-guided random multi-layer perturbation (ERMP) module. This module combines entropy-sensitive gating with stochastic perturbation to adaptively fuse source- and target-dominant features while mitigating domain-specific biases. By injecting randomized updates at multiple feature depths, the ERMP module enhances robustness to domain shifts and promotes stable cross-domain inference under limited-resource settings.

The ERMP module reuses the source-dominant (\(Z_{t \rightarrow s}\)) and target-dominant (\(Z_{s \rightarrow t}\)) features defined earlier and incorporates the entropy terms \(H_{t \rightarrow ts}\) and \(H_{t \rightarrow st}\) computed in the EKD module (Equation~\ref{eq:entropy}). These entropy values guide a gating attention mechanism that adaptively balances the influence of dominant features in refining the target representation \(Z_t\).

A learnable parameter \(\gamma\), modulated by a sigmoid activation \(\sigma(\gamma)\), dynamically controls the relative influence of entropy signals \(H_{t \rightarrow st}\) and \(H_{t \rightarrow ts}\), producing the final attention map:

\begin{equation}
\text{Attn}_{\text{gating}} = (1 - \sigma(\gamma)) \cdot H_{t \rightarrow st} + \sigma(\gamma) \cdot H_{t \rightarrow ts}.
\end{equation}

This dynamic formulation enables the model to adaptively weight the dominant features based on learned uncertainty cues, rather than relying on fixed heuristics.

To enhance robustness, we introduce stochastic perturbation to the target feature. At each training iteration, one of three feature stages—shallow (stage 2), mid (stage 4), or deep (stage 6)—is randomly selected for injection. The updated target feature is then defined as:

\begin{equation}
\tilde{Z}_t = Z_t + \alpha \cdot (\text{Attn}_{\text{gating}} - Z_t),
\end{equation}
where \(\alpha\) controls the perturbation strength. This random injection acts as a multi-level regularizer that encourages the model to rely on stable cross-stage patterns rather than overfitting to specific depths. The combined effect of entropy-aware attention and stochastic perturbation promotes more consistent and transferable feature representations under domain shift.

\section{Experiments}
\label{4}

\subsection{Datasets}
\label{4_1}
We conduct our experiments using the \textbf{Pascal VOC}~\cite{everingham2010pascal}, \textbf{Clipart1k}~\cite{inoue2018cross}, \textbf{Watercolor2k}~\cite{inoue2018cross}, and \textbf{Comic2k}~\cite{inoue2018cross} datasets. We use Pascal VOC2007-trainval and VOC2012-trainval as the source domain, while Clipart1k, Watercolor2k, and Comic2k serve as the target domains. The Pascal VOC dataset~\cite{everingham2010pascal}, which consists of real-world images, includes 16,551 images across 20 distinct object categories. Clipart1k~\cite{inoue2018cross} is a graphic-style dataset with complex backgrounds, containing 1,000 images that match Pascal VOC’s 20 categories. Clipart1k is divided into a training set and a testing set, each containing 500 images. Watercolor2k and Comic2k~\cite{inoue2018cross} each have 2,000 images, split evenly into 1,000 for training and 1,000 for testing. We train on the provided training sets and evaluate on the corresponding test sets.

\subsection{Objective Function}
\label{4_3}
The total loss function in our DA-Mamba model is defined as:

\begin{equation}
L_{\text{total}} = L_{\text{det}} + \lambda_{\text{adv}} L_{\text{adv}} + \lambda_{\text{entropy}} L_{\text{entropy}},
\label{eq:total}
\end{equation}
where \( L_{\text{det}} \) is the detection loss, as described in Equation \ref{eq:det}, consisting of both classification and localization components. The adversarial loss \( L_{\text{adv}} \), detailed in Equation \ref{eq:adv}, promotes domain adaptation by aligning feature distributions across the source and target domains. The knowledge distillation loss \( L_{\text{entropy}} \), described in Equation \ref{eq:kd}, facilitates soft feature alignment between the source-dominant and target-dominant branches. The weighting coefficients \( \lambda_{\text{adv}} \) and \( \lambda_{\text{entropy}} \) are hyperparameters that balance the impact of these auxiliary losses and are specified in the implementation details section.

\subsection{Implementation Details}
\label{4_2}
For all domain adaptation (DA) tasks, we utilize pretrained weights on the ImageNet dataset \cite{deng2009imagenet} as the backbone network in our proposed DA-Mamba method. The model is implemented in two versions: DA-Mamba-S, designed for speed, and DA-Mamba-B, optimized for accuracy. DA-Mamba-S uses $C{=}96$ and layer configuration $\{2, 2, 7, 5, 2, 2\}$, while DA-Mamba-B uses $C{=}128$ and $\{2, 2, 10, 5, 2, 2\}$. Here, $C$ denotes the number of hidden channels in the first stage. The model size, theoretical FLOPs, and inference speed for both versions are reported in Table~\ref{tab:method-comparison}. Optimization is performed using the Stochastic Gradient Descent (SGD) algorithm, with a momentum of 0.9 and a weight decay parameter of $1 \times 10^{-3}$. We employ a base learning rate of $5 \times 10^{-3}$ for the Pascal VOC, WaterColor, Clipart, and Comic2k  datasets. The learning rate follows a warmup cosine scheduler. Across all datasets, the batch size is consistently set to 32, and the model is trained for 100 epochs. The hyperparameters $\lambda_{\text{adv}}$ and $\lambda_{\text{entropy}}$ in the DA-Mamba method are set to $0.5$ and $0.1$, respectively, for all DA tasks, as shown in Equation ~\ref{eq:total}.

\subsection{Experimental Results}
We compare the proposed DA-Mamba with several state-of-the-art DAOD methods, including CNN-based methods such as BSR+WST~\cite{kim2019self}, SWDA~\cite{saito2019strong}, HTCN~\cite{chen2020harmonizing}, I$^3$Net~\cite{chen2021i3net}, DBGL~\cite{chen2021dual}, IDF~\cite{lang2022exploring}, AT~\cite{li2022cross}, LODS~\cite{li2022source}, and CMT~\cite{cao2023contrastive}, as well as Transformer-based method DA-DETR~\cite{zhang2023detr}. The results of SWDA and HTCN are cited from~\cite{chen2021i3net}, and following~\cite{chen2021i3net}, we 
 reproduced the IDF, AT, LODS, and CMT models in our one-stage detection scenario~\cite{wang2024triple}. Source Only denotes that SSD~\cite{liu2016ssd} is trained on the source domain and directly evaluated on the target domain without adaptation.

\textbf{Results on Clipart1K.}  
We compare the proposed DA-Mamba with several state-of-the-art DAOD methods, including CNN-based methods such as TFD~\cite{wang2024triple} and Transformer-based methods such as DA-DETR~\cite{zhang2023detr}. As shown in Table~\ref{table1_clipart}, DA-Mamba-B achieves 45.3\% mAP, surpassing the best CNN-based method TFD (41.2\%) by 4.1\% and the Transformer-based DA-DETR (41.3\%) by 4.0\%. Similarly, DA-Mamba-S achieves 42.1\% mAP, consistently outperforming previous CNN and Transformer-based approaches.

\textbf{Results on Watercolor2K.}  
As reported in Table~\ref{table_watercolor}, DA-Mamba-B achieves 57.8\% mAP, outperforming TFD (55.0\%) by 2.8\% and DA-DETR (50.6\%) by 7.2\%. DA-Mamba-S also improves over previous methods, achieving 54.8\% mAP.

\textbf{Results on Comic2K.}  
Table~\ref{table_comic} shows that DA-Mamba continues to outperform existing methods on Comic2K. DA-Mamba-B achieves 37.1\% mAP compared to 34.4\% for TFD and 35.1\% for DA-DETR. DA-Mamba-S also achieves 35.6\% mAP, demonstrating robust performance across diverse domain shifts.

Compared to CNN-based models, DA-Mamba shows significantly higher performance in categories like Cat, Bird, Dog, Train, and TV across Clipart1k, Watercolor2k, and Comic2k. Against transformer-based models, DA-Mamba achieves superior results in categories such as Car, Person, Bike, and Bird, demonstrating stronger adaptation in both structural and appearance-variant classes.

\begin{figure*}[ht] 
    \centering
    \includegraphics[width=1\linewidth]{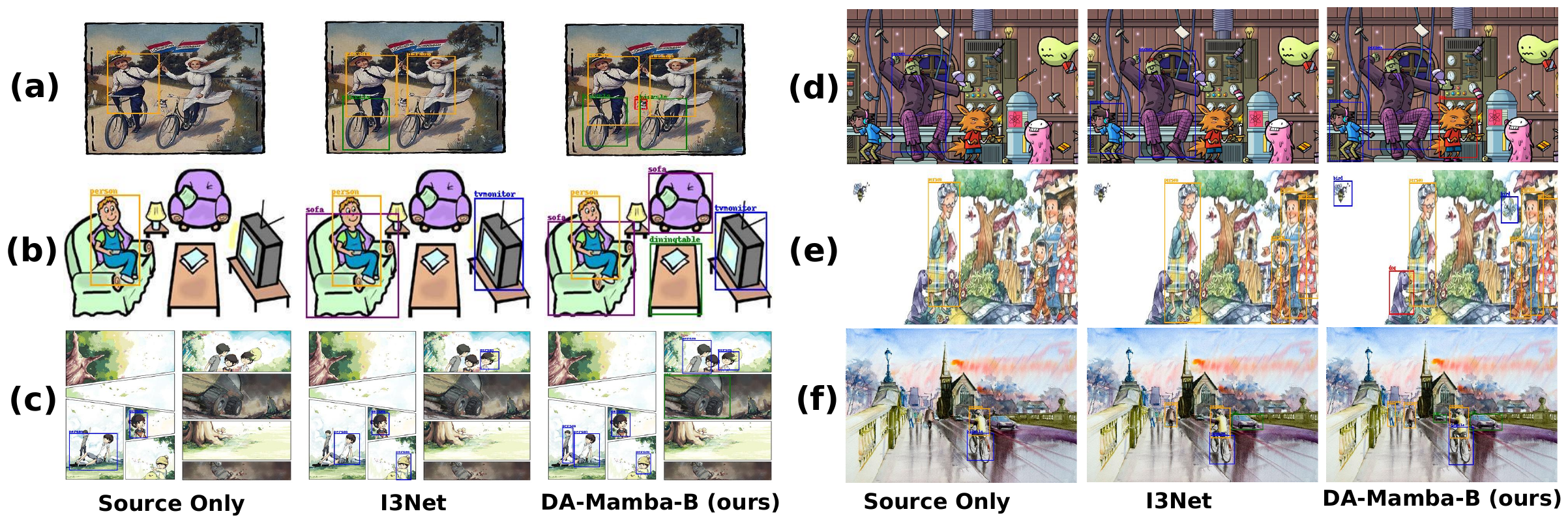}
    \vskip-3.5pt
    \caption{Detection results on Clipart1K ((a), (b)), Comic2K ((c), (d)) and Watercolor2K ((e), (f)) by Source Only~\cite{liu2016ssd}, I3Net~\cite{chen2021i3net} and DA-Mamba-B (Ours).}
    \label{fig:qualitive}
\end{figure*}

\textbf{Qualitative results.}  
The detection results of Source Only~\cite{liu2016ssd}, I$^3$Net~\cite{chen2021i3net}, and DA-Mamba-B on three datasets are displayed in Figure~\ref{fig:qualitive}. The figure shows that DA-Mamba-B outperforms the other detectors in terms of object detection ability. Specifically, DA-Mamba-B not only successfully detects objects missed by conventional methods, but also improves the localization accuracy of bounding boxes, leading to a notable increase in mAP.

\subsection{Model Complexity}
The number of channels and layers used in DA-Mamba-S and DA-Mamba-B are specified in the Implementation Details section, and Table~\ref{tab:method-comparison} provides a comprehensive comparison of their computational complexities during inference.

\begin{table}[H]
    \centering
     \caption{Comparison of methods with parameters, FLOPS and inference time. ”-S” and
            ”-B” indicates that the model is Small and Base, respectively.}
    \scalebox{0.95}{
    \begin{tabular}{l|ccc}
        \toprule
        Method & Params (M) & FLOPs (G) & Inference Time (mS) \\
        \midrule
        DA-Mamba-S  &        62.3      &          8.3   & 30     \\
        DA-Mamba-B  &       91.4      &         12.1  & 48         \\
        \bottomrule
    \end{tabular}
    }
    \label{tab:method-comparison}
\end{table}

\subsection{Ablation Study}
Table \ref{table:abb} highlights the contribution of each component in DA-Mamba-B for PASCAL VOC to Clipart1k adaptation. The Hybrid Domain-Adaptive Mamba-Transformer (HDAMT) blocks play a crucial role, achieving the highest individual performance improvement, demonstrating its effectiveness in learning transferable representations. Additional modules, including Entropy-Based Knowledge Distillation (EKD), Entropy-Guided Random Multi-Layer Perturbation (ERMP), and Adversarial Learning  (ADL), further refine feature alignment, leading to incremental gains.

\begin{table}[H]
\caption{Results on adaptation from Pascal VOC to Clipart1k (\%) using the proposed DA-Mamba-B model.}
\centering
\scalebox{1}{
\begin{tabular}{cccc|c}
\toprule
HDAMT          & EKD          & ERMP & ADL    & mAP        \\ 
\midrule
             &  &   &   &        26.7   \\ 
    \checkmark   &    &   &    &   38.1   \\ 
    & \checkmark    &  &    &   33.4  \\ 
      &      &  \checkmark   &    & 31.1   \\
      &      &     &  \checkmark   & 32.4   \\
      \checkmark & \checkmark     &     &    & 42.1   \\
      \checkmark & \checkmark     &  \checkmark   &    & 43.9   \\
      \checkmark & \checkmark     &  \checkmark   &  \checkmark & \textbf{45.3}   \\
\bottomrule
\end{tabular}
}
\label{table:abb}
\end{table}

\begin{table*}[thb]
\centering
\caption{Results of adapting PASCAL VOC to Clipart1k (\%).  ”-S” and
            ”-B” indicates that the model is Small and Base, respectively.}
\label{table1_clipart}
\renewcommand{\arraystretch}{1.1} 
\setlength\tabcolsep{3.5pt} 
\resizebox{\linewidth}{!}{%
\begin{tabular}{lcccccccccccccccccccccc}
\toprule
\textbf{Methods} & \textbf{Aero} & \textbf{Bicycle} & \textbf{Bird} & \textbf{Boat} & \textbf{Bottle} & \textbf{Bus} & \textbf{Car} & \textbf{Cat} & \textbf{Chair} & \textbf{Cow} & \textbf{Table} & \textbf{Dog} & \textbf{Horse} & \textbf{Bike} & \textbf{Person} & \textbf{Plant} & \textbf{Sheep} & \textbf{Sofa} & \textbf{Train} & \textbf{TV} & \textbf{mAP} \\
\midrule
\multicolumn{22}{c}{\textbf{CNN Based} } \\ 
\midrule         
WST~\cite{kim2019self} & 30.8 & 65.5 & 18.7 & 23.0 & 24.9 & 57.5 & 40.2 & 10.9 & 38.0 & 25.9 & 36.0 & 15.6 & 22.6 & 66.8 & 52.1 & 35.3 & 1.0 & 34.6 & 38.1 & 39.4 & 33.8 \\
BSR~\cite{kim2019self} & 26.3 & 56.8 & 21.9 & 20.0 & 24.7 & 55.3 & 42.9 & 11.4 & 40.5 & 30.5 & 25.7 & 17.3 & 23.2 & 66.9 & 50.9 & 35.2 & 11.0 & 33.2 & 47.1 & 38.7 & 34.0 \\
BSR+WST~\cite{kim2019self} & 28.0 & 64.5 & 23.9 & 19.0 & 21.9 & 64.3 & 43.5 & 16.4 & 42.2 & 25.9 & 30.5 & 7.9 & 25.5 & 67.6 & 54.5 & 36.4 & 10.3 & 31.2 & 57.4 & 43.5 & 35.7 \\
SWDA$^\dagger$~\cite{saito2019strong} & 29.0 & 60.7 & 25.0 & 20.4 & 24.6 & 55.4 & 36.1 & 13.1 & 41.2 & 38.3 & 30.3 & 17.0 & 21.2 & 55.2 & 50.4 & 36.6 & 10.6 & 38.4 & 49.2 & 41.2 & 34.7 \\
HTCN$^\dagger$~\cite{chen2020harmonizing} & 28.7 & 67.7 & 25.3 & 16.1 & 28.7 & 56.0 & 38.9 & 12.5 & 41.0 & 33.0 & 29.6 & 12.9 & 22.9 & 69.0 & 55.9 & 36.1 & 11.8 & 34.1 & 48.8 & 46.8 & 35.8 \\
I$^3$Net~\cite{chen2021i3net} & 30.0 & 67.0 & 32.5 & 21.8 & 29.2 & 62.5 & 41.3 & 11.6 & 37.1 & 39.4 & 27.4 & 19.3 & 25.0 & 67.4 & 55.2 & 42.9 & 19.5 & 36.2 & 50.7 & 39.3 & 37.8 \\
DBGL~\cite{chen2021dual} & 23.2 & 65.5 & 30.1 & 18.3 & 24.6 & 67.6 & 43.9 & 15.1 & 38.7 & 36.4 & 31.3 & 20.2 & 25.0 & 74.3 & 55.1 & 38.2 & 12.5 & 41.0 & 49.1 & 43.9 & 37.7 \\
IDF~\cite{lang2022exploring} & 28.1 & 63.2 & 30.4 & 19.7 & 26.3 & 63.7 & 39.8 & 9.5 & 42.3 & 46.9 & 39.6 & 17.6 & 25.4 & 59.3 & 57.9 & 37.1 & 15.3 & 39.8 & 53.7 & 46.1 & 38.1 \\
AT~\cite{li2022cross} & 30.1 & 68.9 & 29.8 & 27.8 & 28.8 & 67.1 & 45.1 & 14.3 & 43.3 & 47.2 & 35.9 & 20.3 & 30.1 & 65.1 & 57.3 & 37.7 & 13.3 & 38.7 & 43.2 & 42.1 & 39.3 \\
LODS~\cite{li2022source} & 32.1 & 65.1 & 34.7 & 29.9 & \textbf{30.6} & 63.9 & 44.6 & 16.3 & 46.7 & 46.5 & 43.5 & 19.7 & 28.9 & 58.5 & 64.8 & 40.1 & 14.7 & 36.4 & 39.6 & 41.2 & 39.9 \\
CMT~\cite{cao2023contrastive} & 31.9 & 66.9 & 33.9 & 30.2 & 26.3 & 65.2 & 43.6 & 12.6 & 44.5 & 46.3 & 47.9 & 19.3 & 29.9 & 53.1 & 63.3 & 40.5 & 17.1 & 41.2 & 49.6 & 43.9 & 40.4 \\
TFD~\cite{wang2024triple}   & 27.9 & 64.8 & 28.4 & \textbf{29.5} & 25.7 & 64.2 & 47.7 & 13.5 & \textbf{47.5} & 50.9 & \textbf{50.8} & 21.3 & 33.9 & 60.2 & 65.6 & 42.5 & 15.1 & 40.5 & 45.5 & 48.6 & 41.2 \\
\midrule        
\multicolumn{22}{c}{\textbf{Transformer Based} } \\ 
\midrule  
DA-DETR~\cite{zhang2023detr} & \textbf{43.1} & 47.7 & 31.5 & 33.7 & 21.4 & 62.8 & 42.6 & 14.8 & 39.5 & 44.2 & 35.9 & 27.5 & 31.8 & 72.6 & 65.6 & 42.2 & 17.3 & 31.1 & \textbf{71.3} & 50.1 & 41.3 \\
\midrule  
\multicolumn{22}{c}{\textbf{Mamba Based} } \\ 
\midrule        
Source Only~\cite{liu2016ssd} & 27.3 & 60.4 & 17.5 & 16.0 & 14.5 & 43.7 & 32.0 & 10.2 & 38.6 & 15.3 & 24.5 & 16.0 & 18.4 & 49.5 & 30.7 & 30.0 & 2.3 & 23.0 & 35.1 & 29.9 & 26.7 \\
\rowcolor{lightgray!45} \textbf{DA$\-$Mamba-S (Ours)} & 32.3 & 69.0 & 32.3 & 25.0 & 27.2 & 63.7 & 48.1 & 16.5 & 39.2 & 48.1 & 36.3 & 27.4 & 31.6 & 68.5 & 61.4 & 40.9 & 19.2 & 39.7 & 54.7 & 49.1 & 42.1 \\
\rowcolor{lightgray!45} \textbf{DA$\-$Mamba-B (Ours)} & 34.7 & \textbf{74.3} & \textbf{34.8} & 26.9 & 29.2 & \textbf{68.5} & \textbf{51.7} & \textbf{17.7} & 42.2 & \textbf{51.7} & 39.1 & \textbf{29.5} & \textbf{34.0} & \textbf{73.7} & \textbf{66.1} & \textbf{44.0} & \textbf{20.6} & \textbf{42.7} & 58.8 & \textbf{52.8} & \textbf{45.3} \\

\bottomrule
\end{tabular}}
\end{table*}

\begin{table}[htbp]
    \centering
    \caption{Results on adaptation from PASCAL VOC to Watercolor2k (\%). ”-S” and
            ”-B” indicates that the model is Small and Base, respectively.}
    \label{table_watercolor}
    \small
    \setlength\tabcolsep{4.5pt} 
    \renewcommand{\arraystretch}{1.1} 
    \resizebox{1\linewidth}{!}{%
    \begin{tabular}{lccccccc}
        \toprule
        \textbf{Methods} & \textbf{Bike} & \textbf{Bird} & \textbf{Car} & \textbf{Cat} & \textbf{Dog} & \textbf{Person} & \textbf{mAP} \\
        \midrule
        \multicolumn{8}{c}{\textbf{CNN Based} } \\ 
        \midrule  
        BSR~\cite{kim2019self} & 82.8 & 43.2 & 49.8 & 29.6 & 27.6 & 58.4 & 48.6 \\
        WST~\cite{kim2019self} & 77.8 & 48.0 & 45.2 & 30.4 & 29.5 & 64.2 & 49.2 \\
        SWDA$^\dagger$~\cite{saito2019strong} & 73.9 & 48.6 & 44.3 & 36.2 & 31.7 & 62.1 & 49.5 \\
        BSR+WST~\cite{kim2019self} & 75.6 & 45.8 & 49.3 & 34.1 & 30.3 & 64.1 & 49.9 \\
        HTCN$^\dagger$~\cite{chen2020harmonizing} & 78.6 & 47.5 & 45.6 & 35.4 & 31.0 & 62.2 & 50.1 \\
        I$^3$Net~\cite{chen2021i3net} & 81.1 & 49.3 & 46.2 & 35.0 & 31.9 & 65.7 & 51.5 \\
        DBGL~\cite{chen2021dual} & 84.0 & 46.7 & 45.5 & 36.2 & 35.7 & 63.7 & 52.0 \\
        IDF~\cite{lang2022exploring} & 86.1 & 45.6 & 47.8 & 35.7 & 34.2 & 63.3 & 52.1 \\
        AT~\cite{li2022cross} & 85.8 & 49.6 & 48.3 & 37.8 & 32.5 & 63.3 & 52.9 \\
        LODS~\cite{li2022source} & 84.1 & 50.5 & 48.7 & 38.1 & 31.4 & 65.2 & 53.0 \\
        CMT~\cite{cao2023contrastive}  & 87.1 & 48.7 & 50.2 & 37.1 & 31.5 & 66.3 & 53.5 \\
        TFD~\cite{wang2024triple}  & \textbf{93.0} & 52.6 & 47.6 & 39.2 & 33.7 & 63.9 & 55.0 \\
        \midrule  
        \multicolumn{8}{c}{\textbf{Transformer Based} } \\ 
        \midrule  
        DA-DETR~\cite{zhang2023detr} & 58.6 & 53.7 & 31.9 & \textbf{46.2} & \textbf{40.2} & 73.0 & 50.6 \\
        \midrule  
        \multicolumn{8}{c}{\textbf{Mamba Based} } \\ 
        \midrule  
        Source Only~\cite{liu2016ssd} & 77.5 & 46.1 & 44.6 & 30.0 & 26.0 & 58.6 & 47.1 \\
        \rowcolor{lightgray!45} \textbf{DA$\-$Mamba-S (Ours)} & 85.2 & 53.0 & 51.4 & 34.7 & 30.5 & 70.1 & 54.8 \\
\rowcolor{lightgray!45} \textbf{DA$\-$Mamba-B (Ours)} & 88.9 & \textbf{55.2} & \textbf{52.8} & 36.2 & 31.7 & \textbf{73.4} & \textbf{57.8} \\
        \bottomrule
    \end{tabular}}
\end{table}

\begin{table}[htbp]
    \centering
    \caption{Results on adaptation from PASCAL VOC to Comic2K (\%). ”-S” and
            ”-B” indicates that the model is Small and Base, respectively.}
    \label{table_comic}
    \small
    \setlength\tabcolsep{4.5pt} 
    \renewcommand{\arraystretch}{1.1} 
    \resizebox{1\linewidth}{!}{%
    \begin{tabular}{lccccccc}
        \toprule
        \textbf{Methods} & \textbf{Bike} & \textbf{Bird} & \textbf{Car} & \textbf{Cat} & \textbf{Dog} & \textbf{Person} & \textbf{mAP} \\
        \midrule
        \multicolumn{8}{c}{\textbf{CNN Based} } \\ 
        \midrule  
        WST~\cite{kim2019self} & 45.7 & 9.3 & 30.4 & 9.1 & 10.9 & 46.9 & 25.4 \\
        BSR~\cite{kim2019self}  & 45.2 & 15.8 & 26.3 & 9.9 & 15.8 & 39.7 & 25.5 \\
        BSR+WST~\cite{kim2019self} & 50.6 & 13.6 & 31.0 & 7.5 & 16.4 & 41.4 & 26.8 \\
        SWDA$^\dagger$~\cite{saito2019strong}  & 47.4 & 12.9 & 29.5 & 12.7 & 19.1 & 44.1 & 27.6 \\
        HTCN$^\dagger$~\cite{chen2020harmonizing} & 50.3 & 15.0 & 27.1 & 9.4 & 18.9 & 46.2 & 27.8 \\
        I$^3$Net~\cite{chen2021i3net} & 47.5 & 19.9 & 33.2 & 11.4 & 19.4 & 49.1 & 30.1 \\
        DBGL~\cite{chen2021dual} & 45.4 & 15.9 & 24.8 & 11.5 & 29.4 & 55.1 & 30.4 \\
        IDF~\cite{lang2022exploring} & 46.7 & 18.5 & 31.1 & 15.5 & 25.7 & 48.6 & 31.0 \\
        AT~\cite{li2022cross} & 49.4 & 18.3 & 32.1 & 15.2 & 28.7 & 50.3 & 32.3 \\
        LODS~\cite{li2022source} & 50.6 & 18.6 & 31.4 & 13.9 & 28.9 & 49.8 & 32.2 \\
        CMT~\cite{cao2023contrastive}  & 49.8 & 19.2 & 29.8 & 15.2 & 29.1 & 54.1 & 32.9 \\
        TFD~\cite{wang2024triple} & 53.4 & 19.2 & 35.0 & 16.1 & 33.2 & 49.2 & 34.4 \\
        \midrule  
        \multicolumn{8}{c}{\textbf{Transformer Based} } \\ 
        \midrule  
        DA-DETR~\cite{zhang2023detr} & 44.2 & 18.1 & 25.0 & \textbf{27.7} & 33.0 & \textbf{62.4} & 35.1 \\
        \midrule  
        \multicolumn{8}{c}{\textbf{Mamba Based} } \\ 
        \midrule  
        Source Only & 43.3 & 9.4 & 23.6 & 9.8 & 10.9 & 34.2 & 21.9 \\
        \rowcolor{lightgray!45} \textbf{DA$\-$Mamba-S (Ours)} & 55.2 & 20.1 & 36.1 & 16.8 & 34.5 & 50.3 & 35.6 \\
        \rowcolor{lightgray!45} \textbf{DA$\-$Mamba-B (Ours)} & \textbf{57.0} & \textbf{21.4} & \textbf{37.8} & 17.5 & \textbf{36.0} & 52.0 & \textbf{37.1} \\
        \bottomrule
    \end{tabular}}
\end{table}

\subsection{Analysis of Source and Target Dominant Feature Effect}
To assess the impact of using Source-Target Dominant (STD) features generated via cross-attention (i.e $Z_{s \rightarrow t}$ and $Z_{t \rightarrow s}$), we conduct an ablation study under the PASCAL VOC to Clipart1k adaptation setting using our DA-Mamba-B model. As shown in Table \ref{table:std}, replacing Source-Target (ST) features with 
 Source-Target-Dominant (STD) features results in a notable improvement, increasing mAP from 42.6\% to 45.3\%. This performance boost highlights the limitations of ST features, which tend to retain rigid, non-transferable domain-specific characteristics, leading to suboptimal adaptation. In contrast, STD features effectively suppress domain biases, facilitating a softer, more adaptable feature representation. 

\begin{table}[H]
\caption{Analysis of ST and STD feature effects. ST denotes Source-Target features ($Z_{s}$ and $Z_{t}$), while STD denotes source-target dominant features ($Z_{s \rightarrow t}$ and $Z_{t \rightarrow s}$).}
\centering
\scalebox{1.1}{
\begin{tabular}{ccc|c}
\hline
EKD          & ERMP & ADL    & mAP        \\ 
\hline 
      ST &  ST  &  ST &  42.6         \\ 
      STD & STD & STD  &  \textbf{45.3}  \\
    
\hline
\end{tabular}
}
\label{table:std}
\end{table}

\subsection{Analysis of Entropy Knowledge Distillation}
\begin{figure}[H] 
    \centering
    \includegraphics[width=0.85\linewidth]{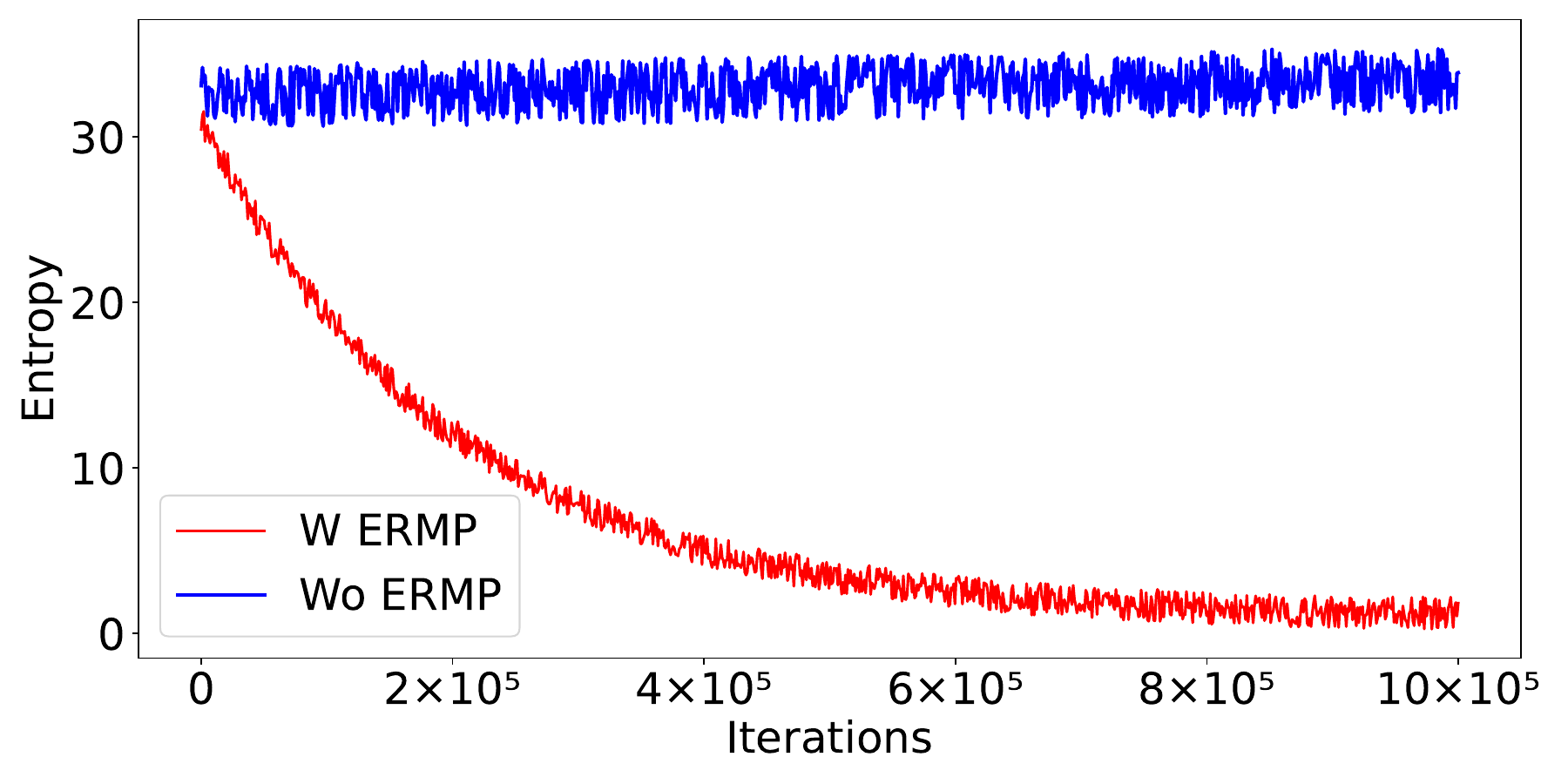}
    \vskip-3.5pt
    \caption{Total entropy of shallow, mid, and deep feature levels measured on adaptation from Pascal VOC to Clipart1k (\%) using the proposed DA-Mamba-B model.}
    \label{fig:entropy}
\end{figure}

Figure~\ref{fig:entropy} presents the entropy loss curves computed across shallow, mid, and deep feature levels, as defined in Equation~\ref{eq:kd} under the PASCAL VOC to Clipart1k adaptation setting using our DA-Mamba-B model. It can be observed that the ERMP module consistently reduces entropy, indicating enhanced feature confidence and improved discrimination between domains. This reduction in entropy suggests that ERMP facilitates more effective transfer of domain-invariant features by suppressing noisy activations and minimizing uncertainty during knowledge distillation. The effectiveness of the ERMP module is validated by the adaptation performance gains reported in Table~\ref{table:abb}.

\section{Conclusion}
We proposed DA-Mamba, the first domain-adaptive object detection framework based on Mamba, integrating efficient state-space modeling with Transformer-based attention within a single-stage architecture. By employing adaptive feature scanning, entropy-guided knowledge distillation, and entropy-sensitive feature fusion, DA-Mamba achieves robust cross-domain feature alignment while mitigating overfitting and suppressing noisy domain-specific activations. Extensive evaluations demonstrate that DA-Mamba achieves state-of-the-art performance on benchmark UDA datasets, combining strong generalization with high computational efficiency.

\bibliographystyle{ieeetr}
\bibliography{damamba}

\end{document}